\documentclass[runningheads]{llncs}

\usepackage{todonotes}

\usepackage{eccv}

\usepackage{eccvabbrv}

\usepackage{makecell}
\usepackage{graphicx}
\usepackage{booktabs}
\usepackage{listings}
\usepackage{caption}
\usepackage{dirtree}
\DeclareCaptionType{listing}[Listing][List of Listings]
\usepackage[accsupp]{axessibility}  
\usepackage{tabularx}

\usepackage{lipsum}
\usepackage{siunitx}
\DeclareSIUnit\px{px}
\DeclareSIUnit\fps{fps}

\usepackage[acronym]{glossaries}

\newacronym{uav}{UAV}{Unmanned Aerial Vehicle}
\newacronym{pot}{POT}{Pixels on Target}
\newacronym{fov}{FOV}{Field of View}
\newacronym{ml}{ML}{Machine Learning}
\newacronym{ir}{IR}{Infra-red}
\newacronym{erc}{ERC}{Event Rate Control}
\newacronym{sod}{SOD}{Small Object Detection}
\newacronym{fpn}{FPN}{Feature Pyramid Network}

\usepackage{hyperref}

\usepackage{orcidlink}

\begin{document}

\title{SkyEV: RGB-Event UAV detection and tracking dataset and baseline} 

\titlerunning{SkyEV: RGB-Event Dataset}

\author{Jakub Mandula\inst{1}\orcidlink{0009-0009-4894-6800} \and
Sebastian Heusinger\inst{2}\orcidlink{0009-0009-4894-6800} \and
Julian Moosmann\inst{1}\orcidlink{0009-0009-4894-6800} \and
Christian Vogt\inst{1}\orcidlink{0000-0003-4551-4876} \and
Michele Magno\inst{1}\orcidlink{0000-0003-0368-8923}}

\authorrunning{J.~Mandula et al.}

\institute{ETH Zurich \\
\email{\{jakub.mandula,christian.vogt,michele.magno\}@pbl.ee.ethz.ch} \and
\email{sheusinger@ethz.ch}}
\maketitle

\begin{abstract}
Detecting \Glspl{uav} in air spaces has become increasingly important due to \Glspl{uav} widespread availability and easy usage. However, due to their small size, they are typically difficult to detect at a sufficient range. For the training of optimized detection algorithms, datasets have been published, covering optical sensing methods ranging from infrared to regular RGB to event-sensor-based. However, these datasets often fail to reflect realistic counter-UAV scenarios, lacking critical factors such as camera ego-motion, extremely small target scales, and diverse lens configurations, and introduce compression artefacts on the frame images. To address this gap, we introduce SkyEV, an open-source dataset featuring highly synchronized uncompressed RGB and event-based data. SkyEV distinguishes itself by capturing complex real-world conditions, including significant camera motion and varied optical setups, which are essential for testing the fundamental trade-off between Field of View and detection range. Furthermore, we provide a unified data loader and establish an experimental baseline using a multi-modal architecture, demonstrating the dataset's efficacy in detecting challenging, small-scale targets. 

  \keywords{UAV Detection \and Event-based Vision \and Multi-modal Dataset \and Object Tracking \and Sensor Fusion}
\end{abstract}

\section{Introduction}
\label{sec:intro}

As highlighted by recent European research, detecting \glspl{uav} near sensitive infrastructure, such as airports or other no-fly zones, has become a critical priority due to the rising number of security incidents \cite{2024stepniakResearchInnovationDrones}. Accordingly, there is growing research interest, shown by more than 550 studies being published on drone detection at airports between 2014 and 2025 \cite{de_macedo_drone_2025}.

State-of-the-art detection methods are categorized as either active or passive. Active systems, such as RADAR \cite{2026khawajaSurveyDetectionClassification} or LiDAR \cite{2018hammerLidarbasedDetectionTracking}, provide precise localization but require emitting a sensing signal. Passive methods rely on signals emitted by the \gls{uav} itself, such as radio frequency emissions, noise, heat, or visible light \cite{2025dongSecuringSkiesComprehensive}. While passive radio and acoustic techniques struggle with precise localization \cite{2026jekaterynczukSoundUAVsReview}, vision-based sensing (infrared or visible spectrum) enables accurate identification and localization, making it a prominent focus for medium-range detection\cite{2022zhaoVisionBasedAntiUAVDetection, 2025dingVisionBasedUnmannedAerial,2025dongSecuringSkiesComprehensive}.
\begin{figure}[t]
    \centering
    \includegraphics{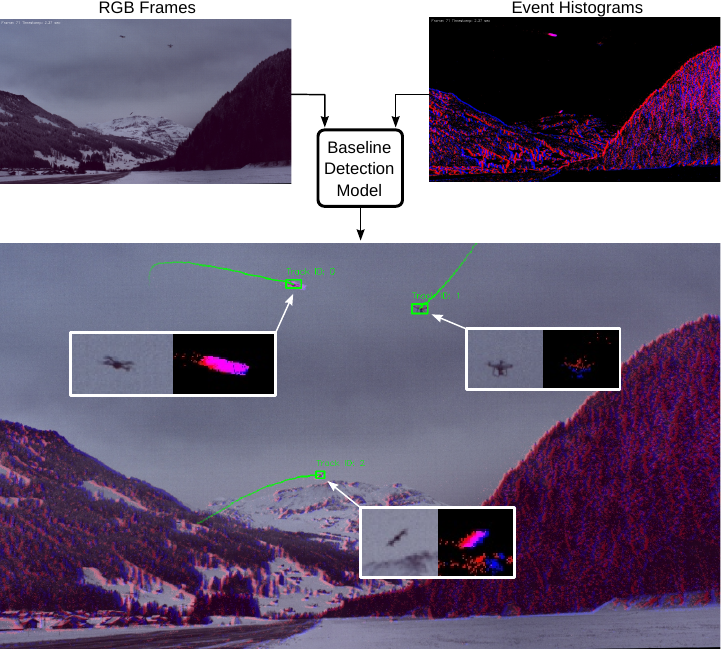}
    \caption{Example image from the SkyEV dataset containing event data overlayed on top of an RGB frame with multiple annotated UAV tracks. Ego-motion of the camera causes background clutter also in the event domain.}
    \label{fig:example_drone_tracking}
\end{figure}

Despite being a passive and relatively low-cost solution, traditional optical \gls{uav} detection suffers from several challenges: limited contrast and dynamic range due to camera sensors with fixed shutter times. This limits detection effectiveness, especially in high-contrast scenarios, like \glspl{uav} flying close to the sun or against a cluttered background. High velocity of the flying objects, as well as the necessity to cover a large field of view, requires robustness to motion blur and a relatively high frame rate to resolve these small objects at a long range \cite{magriniDroneDetectionEvent}. These challenges are partially mitigated by novel event-based cameras, providing a sensor technology with intrinsic properties particularly advantageous for \gls{uav} detection. These sensors allow asynchronous capture of any brightness changes in the scene, effectively adjusting brightness on a pixel level. This differential mode of operation results in an exceedingly high dynamic range $>$\SI{120}{\dB} and temporal resolutions equivalent to \SI{10000}{\fps} \cite{2008lichtsteiner128times128120,2022gallegoEventBasedVisionSurvey,2022gallegoEventBasedVisionSurvey}. Additionally, their inherently sparse data stream significantly accelerates processing time \cite{SAST}.

While these event-based sensors provide exceptional speed and dynamic range, event-based sensors have distinct limitations: their reliance on contrast changes complicates the detection of slow-moving targets \cite{2022gallegoEventBasedVisionSurvey}. They also suffer from relatively low spatial resolution (e.g., max. 1280$\times$800) \cite{ 2022gallegoEventBasedVisionSurvey,2024chakravarthiRecentEventCamera}. Consequently, pairing these novel sensors with classical frame cameras is recognized as essential in improving the robustness of \gls{uav} detection  \cite{2025magriniFREDFlorenceRGBEvent, 2025wangLongTermVisualObject,2024magriniNeuromorphicDroneDetection}. 

Within \gls{ml} based optical detection, tracking of \gls{uav} poses unique challenges that distinguish it from general object detection tasks. A primary hurdle is the size of a \gls{uav} in pixels on the camera sensor. Optical systems require sufficient well-exposed \gls{pot} to resolve critical spatial features for confident object detection and classification \cite{2022emersonTopologicalApproachMotion}. This requirement imposes a fundamental triangular trade-off between detection range, \gls{fov}, and system latency at a given data bandwidth and compute power. For example, extending the detection range while maintaining a wide \gls{fov} necessitates high-resolution sensors, which exponentially increase data bandwidth and computational load. To reliably train \gls{ml} algorithms, datasets must reflect these trade-offs by benchmarking tracking accuracy across large, as well as relatively small, bounding boxes (spatial dimension \(\sqrt{w\times h} < \SI{20}{\px}\)). Another attribute specific to \gls{uav} detection is their velocity. \gls{uav} can exhibit high relative velocities (\(>\SI{1000}{\px\per\second}\)) or stay near stationary from the point of view, and traverse complex, non-homogeneous backgrounds such as urban areas, forests, or cloudy skies, making detection a challenge \cite{2025dingVisionBasedUnmannedAerial}. 

Keeping these complex environments, behavior, and varying illumination situations in mind, \gls{ml} algorithm training requires a large amount of high-quality training data, as well as efficient and small \gls{ml} models for real-time detection of \gls{uav}. With this paper, we strive to contribute the following: 
\begin{enumerate}
\item We provide the open-source SkyEV dataset comprising 2 synchronized event cameras and one RGB camera running at 30 and 60 FPS, offering recordings of 8 different drone types with an average pixel size of 83x49 pixels and a total sequence length of \qty{2.17}{\hour} traversing complex environment and light scenarios. 
\item In addition, we present a baseline state-of-the-art algorithm fusing event-based and frame-based camera images for robust drone detection to demonstrate the benefit of utilizing multi-modal source data. Its capability for future fast inference on small, embedded systems is enhanced by directly embedding an SAST \cite{SAST} model to filter promising image patches from the event camera histograms, allowing the sparse processing of potentially relevant patches.
\item We present the baseline algorithm's performance on the SkyEV dataset.
\end{enumerate}
\pagebreak
\section{Related Work}
\label{sec:related_work}
Despite the rapid growth of event-based vision \cite{magriniDroneDetectionEvent}, the domain is characterized by a disparity between the abundance of processing methods and the scarcity of diverse datasets \cite{2024chakravarthiRecentEventCamera}. Current benchmarks are dominated by automotive-centric datasets, such as the DSEC \cite{2021gehrigDSECStereoEvent}, a 53-minute stereo dataset focusing on depth perception and optical flow, or the 1Mpx dataset \cite{2020perotLearningDetectObjects}, an extensive 14.7h detection dataset with bounding box object labels. In contrast, the unique constraints of \gls{uav} detection - including small-target and high velocity, necessitate a departure from automotive-centered benchmark datasets in favor of more domain-specific resources. 

\subsection{Drone-detection datasets} 
Several recent works have focused on the topic of optical drone detection, specifically utilizing frame-based, \gls{ir}, and event-based cameras \cite{2022zhaoVisionBasedAntiUAVDetection,2025dingVisionBasedUnmannedAerial,magriniDroneDetectionEvent}. Summarized in \cref{tab:uav_datasets} are current state-of-the-art datasets that provide bounding-box or tracklet-based labels for sequences in the \gls{uav} detection task. We compare them by the following key performance indicators where available: sensor types (RGB, Event, IR) and resolution, number of recorded frames, average scale of the bounding boxes, frames per second (FPS), duration of the complete dataset, ratio of UAV sequences in the complete dataset, number of drone types and general number of classes if the camera has self-motion or is static and finally if the dataset is publically available.

\begin{table*}[ht]
\centering
\caption{Comprehensive overview of UAV detection and tracking datasets with frame RGB, IR, and asynchronous event data.}
\label{tab:uav_datasets}
\resizebox{\textwidth}{!}{
\begin{tabular}{lcccccccccccccc}
\toprule
\textbf{Dataset} & \textbf{Year} & \textbf{Resolution} & \textbf{RGB} & \textbf{Event} & \textbf{IR} & \textbf{\makecell{\# \\ Frames}} & \textbf{AVG Scale ($\sqrt{wh}$)} & \textbf{FPS} & \textbf{Duration} & \textbf{\makecell{Drone \\ Types}} &  \textbf{\makecell{UAV \\ Seq. Ratio}} & \textbf{\makecell{Camera \\ Motion}} & \textbf{Classes} & \textbf{Public}\\
\midrule
Anti-UAV300~\cite{2023jiangAntiUAVLargeScaleBenchmark} & 2023 & 640$\times$512 & \checkmark & $\times$   & \checkmark & 585,900 &  ($\sim$50)                    & 25  & 6h:30m & $>$6  & 100   & \checkmark & 1 & \checkmark  \\
Anti-UAV600~\cite{2023zhuEvidentialDetectionTracking} & 2023 & 640$\times$512                     & $\times$   & $\times$   & \checkmark & 723,000 & 30$\times$19 (23.8)                  & 25  & 8h:01m & $>$6 & 100   & \checkmark&   1 & \checkmark  \\
CRSOT~\cite{2024zhuCRSOTCrossResolutionObject}             & 2023 &  1280$\times$720                    & \checkmark & \checkmark & $\times$   & 304,974       &    -                 & 25  & 2h:50m & 1     & 100   & \checkmark          & 1  & \checkmark\\
VisEvent~\cite{2023wangVisEventReliableObject}       & 2023 & 346$\times$260                     & \checkmark & \checkmark & $\times$   & 371,127 & 84$\times$66         & 25  & 3h:24m & 1           & 15.97 & \checkmark & -  & \checkmark\\
EventVOT~\cite{2023wangEventStreambasedVisual}       & 2024 & 1280$\times$720                    & $\times$   & \checkmark & $\times$   & 569,359 & 129$\times$100 (113.6)     & -   & -      & -     & 8.41  & \checkmark          & 19 & \checkmark\\
EvDET200K~\cite{2024wangObjectDetectionUsing}       & 2024 & 1280$\times$720                    & $\times$   & \checkmark & $\times$   & 202,260 & 64$\times$72 (67)     & -   & -      & -    &  <1 & \checkmark           & 10 & \checkmark\\
F-UAV-D~\cite{2024mandulaRealTimeFastUnmanned}         & 2024 & 1280$\times$720                    & \checkmark & \checkmark & $\times$   & -       & -                    & 30  & 0h:30m & 2    &  100   & \checkmark & 1  & \checkmark\\
NeRDD~\cite{2024magriniNeuromorphicDroneDetection}             & 2024 & 1280$\times$720                    & \checkmark & \checkmark & $\times$   & 417,497       & 55$\times$31         & 30  & 3h:30m & 2     & 100   & $\times$           & 1 & \checkmark \\
RGBT-Tiny~\cite{2025yingVisibleThermalTinyObject}             & 2025 & 640$\times$512                    & \checkmark &  $\times$ & \checkmark    & 93,000       & -        & 15  & - & -    & 2   & $\times$           & 1 & $\times$ \\
LRDDv2~\cite{2025rouhiLRDDv2EnhancedLongRange}             & 2025 & 1920$\times$1080                    & \checkmark & $\times$ & $\times$   &39,516      &  -       & 30  & - & 2     & 100  & \checkmark            & 1  & $\times$\\
FELT~\cite{2025wangLongTermVisualObject}             & 2025 & 346$\times$260                    & \checkmark & \checkmark & $\times$   &1,949,680       &  -       & -  & - & -     & -   & \checkmark            & 60  & $\times$\\
Ev-Flying~\cite{2025magriniEVFlyingEventbasedDataset}     & 2025 & 1280$\times$720                    & $\times$   & \checkmark & $\times$   & -       &  -                    & 30   & 1h:07m & 1    & 100   & $\times$            & 1  & $\times$ \\
Ev-UAV~\cite{2025chenEventbasedTinyObject}           & 2025 & 346$\times$260                     & $\times$   & \checkmark & $\times$   & -       & 6.8$\times$5.4 (6) & -   & 0h:15m & -    & 100   & $\times$           & 1  & \checkmark\\
CST Anti-UAV~\cite{2025xieCSTAntiUAVThermal}& 2025 & 640$\times$512                     & $\times$   & $\times$   & \checkmark & 240,000 &  ($\sim$18)                    & 25  & 2h:40m & -    &  100   & \checkmark           & 1 & \checkmark \\
FRED~\cite{2025magriniFREDFlorenceRGBEvent}               & 2025 & 1280$\times$720                    & \checkmark & \checkmark & $\times$   & 808,953      &  54.4$\times$34.4 (46.8)                    & 30  & 7h:07m & 5    & 100   & $\times$          & 1  & \checkmark\\
\textbf{SkyEV (ours)} & \textbf{2026} & \textbf{1280$\times$720 } & \textbf{\checkmark} & \textbf{\checkmark }& $\times$   & \textbf{242,465}       & \textbf{83$\times$49  (63.8)  }             & \textbf{30+60}  &\textbf{ 2h:10m }& \textbf{8}    & \textbf{100}   & \textbf{\checkmark}           & \textbf{1} & \checkmark \\
\bottomrule
\end{tabular}
  }
\end{table*}

Initial efforts in event-based datasets for general object detection and tracking, such as VisEvent \cite{2023wangVisEventReliableObject} and EventVOT \cite{2023wangEventStreambasedVisual}, provided valuable foundations for training and evaluating drone-detection algorithms. Both datasets include extensive camera ego-motion, background clutter, and varying lighting conditions. However, they are not strictly drone-centric, and \gls{uav} sequences represent with 16\% and 8.4\% respectively, a small fraction of their volume. 
Consequently, they fail to capture the unique spatial and temporal challenges of \gls{uav} detection outlined in \cref{sec:intro}, such as small target size and rapid velocity changes against high-contrast sky backgrounds. 

To address the specific need for counter-\gls{uav} scenarios, several dedicated datasets have emerged. Datasets like Anti-UAV300 \cite{2023jiangAntiUAVLargeScaleBenchmark}, Anti-UAV600 \cite{2023zhuEvidentialDetectionTracking}, RGBT-Tiny \cite{2023wangVisEventReliableObject} and CST Anti-UAV \cite{2025xieCSTAntiUAVThermal} offer extensive \gls{uav} tracking sequences (at 640$\times$512 resolution). However, they rely entirely on conventional frame-based and \gls{ir} modalities, lacking the high-dynamic-range and high-speed advantages of event-cameras. Conversely, datasets such as Ev-Flying \cite{2025magriniEVFlyingEventbasedDataset}, EvDET200K \cite{2024wangObjectDetectionUsing}, and EV-UAV \cite{2025chenEventbasedTinyObject} use event cameras but lack synchronized frame-based data, precluding the complementary multi-modal approach necessary for robust detection of both fast and near-stationary targets \cite{2023zhouRGBEventFusionMoving,2025bonazziRGBEventFusionSelfAttention,2024magriniNeuromorphicDroneDetection}. In addition, EV-UAV, as well as FELT \cite{2025wangLongTermVisualObject} and VisEvent \cite{2023wangVisEventReliableObject}, use a relatively old 346$\times$260 low-resolution event cameras resulting in limited resolving range.

Among the most recent works, CRSOT \cite{2024zhuCRSOTCrossResolutionObject}, F-UAV-D \cite{2024mandulaRealTimeFastUnmanned}, and FRED \cite{2025magriniFREDFlorenceRGBEvent}, along with subsets NeRDD \cite{2024magriniNeuromorphicDroneDetection}, can serve as valuable inspirations for a robust multi-modal dataset. FRED, for instance, provides an extensive dataset with over 7 hours of varied scenarios. However, all the sequences are filmed with a stationary camera, resulting in no background clutter in the event domain. Furthermore, only about  0.4\% of the bounding boxes have velocities \(>\SI{1000}{\px\per\second}\) and the maximum recorded velocity is around \SI{2200}{\px\per\second}.

Most of these multi-modal datasets ensure some level of spatial and temporal synchronization either through hardware timestamping and camera calibration or through manual alignment. Although there are some methods \cite{2024zhuCRSOTCrossResolutionObject, 2025zhangFrequencyAdaptiveLowLatencyObject} that can handle misaligned multi-modal data, it is generally considered important to keep the events and frames in sync. To this end, multiple methods are proposed in the literature \cite{2023moosmannQuantitativeEvaluationMultiModal, 2021gehrigDSECStereoEvent, 2024pengCoSECCoaxialStereo}.

For spatial alignment, some works utilize Coaxial cameras with the help of a beam splitter. This does allow for the complete elimination of the parallax issue, however effectively halves the light quantity reaching the sensor \cite{2024pengCoSECCoaxialStereo}. Another method relies on the Davis346 camera, containing both event and frame pixels registered on the sensor. However, the Davis346 only contains a low-resolution 346$\times$260 sensor, inadequate for long-range object detection \cite{2022gallegoEventBasedVisionSurvey}. The final alternative is using co-aligned frame and event cameras with minimal baseline and performing a homography transformation of the view \cite{2023jiangAntiUAVLargeScaleBenchmark,2024zhuCRSOTCrossResolutionObject,2025magriniFREDFlorenceRGBEvent,2024magriniNeuromorphicDroneDetection,2024mandulaRealTimeFastUnmanned, 2025xieCSTAntiUAVThermal,2021gehrigDSECStereoEvent}. 

In summary, a critical dissection of these multi-modal datasets reveals several persisting gaps that limit their applicability to real-world deployment:
\begin{itemize}
\item \textbf{Lack of Raw Image data:} While most datasets provide a raw form of event data ($E = \{e_i\}_{i=1}^N = \{(x_i, y_i, t_i, p_i)\}_{i=1}^N$), the frame-based video data is often recorded in compressed form (JPEG, MP4). For certain evaluation tasks and accurate dynamic range comparison, the frames should be recorded with their full bit-depth and without compression artifacts \cite{2025duanEventAidBenchmarkingEventAideda}.
\item \textbf{Lack of Camera Motion:} Event-based sensors trigger on brightness changes. While stationary cameras allow for clean capture of a moving \gls{uav}, real-world counter-\gls{uav} systems (e.g., pan-tilt-zoom trackers or camera-equipped interceptor drones) require a moving viewpoint. Camera ego-motion induces massive amounts of background events \cite{2023moosmannQuantitativeEvaluationMultiModal}, drastically increasing the difficulty of target isolation \cite{2025steinhauserDataQualityMatters} — a challenge largely underrepresented in current datasets. 
\item \textbf{Limited Optical Variance:} Existing datasets primarily utilize fixed lens configurations and sample objects at close range. They fail to capture the fundamental trade-off between \gls{fov} and detection range, lacking the varied image scales required to train models capable of detecting targets at vast distances (where the spatial dimension drops well below $20\times 20$ pixels) \cite{2025dingVisionBasedUnmannedAerial}.
\item \textbf{Benchmarking Friction:} Finally, the lack of a standardized, unified data loader across these datasets complicates the training and fair evaluation of novel \gls{ml} architectures.
\end{itemize}
To overcome these limitations, our proposed SkyEV dataset transitions from idealized recording conditions to realistic, highly demanding counter-\gls{uav} scenarios. By providing synchronized RGB-Event data with significant camera motion, varied lens configurations, and a unified data loader, we establish a robust benchmark for the next generation of multi-modal small-object detection algorithms.

\subsection{Event-based Object detection}
Effective drone detection necessitates architectures that balance the high semantic density of RGB frames with the high temporal resolution and sparsity of event data. Current state-of-the-art event-based detectors, such as the Recurrent Vision Transformer (RVT) \cite{10204090} and the Scene Adaptive Sparse Transformer (SAST) \cite{SAST}, have established high-performance benchmarks by leveraging recurrent temporal blocks and sparse token-based filtering to process event streams efficiently. In parallel, CNN-based detection methods such as Faster-RCNN \cite{2017chenDeepLearningApproacha} or YOLO \cite{2026sapkotaYOLO26KeyArchitectural, 2020yuanLowAltitudeSmall,2021geYOLOXExceedingYOLO} remain the gold standard for high-speed, high-accuracy detection in the RGB domain due to their robust feature extraction capabilities. Recently, transformer-based models such as RT-DETR \cite{2024zhaoDETRsBeatYOLOs} started catching up in terms of accuracy and inference speed, providing accurate object detection even at large resolutions \cite{2021liuSwinTransformerHierarchical}. Promising better scaling with large quantities of data \cite{2020dosovitskiyImageWorth16x16}, as well as easier fusion across other modalities \cite{2021radfordLearningTransferableVisual}.

While these models excel in their respective modalities, recent work such as NeRDD \cite{2024magriniNeuromorphicDroneDetection} suggests that the most resilient detection performance is achieved through multimodal fusion, which compensates for the limitations of individual sensors \cite{2025bonazziRGBEventFusionSelfAttention}. Our work builds on these foundations by integrating a modified SAST module with a YOLOX backbone, thereby aligning and fusing the sparse, motion-centric features of event data with the dense semantic information of RGB frames to provide a comprehensive detection baseline \cite{2025yangFocusMotionRGBEvent}.

\section{SkyEV Dataset}
In this section, we describe the process of recording synchronized event data and RGB frames and the semi-automatic labeling protocol inspired by \cite{2020perotLearningDetectObjects}  used to provide accurate tracking labels for the dataset. 

\begin{figure}[htbp!]
\vspace{-1em}

     \centering
     \begin{subfigure}[b]{0.16\textwidth}
         \centering
         \caption*{Clutter}
         \includegraphics[width=\textwidth, height=\textwidth, keepaspectratio=false]{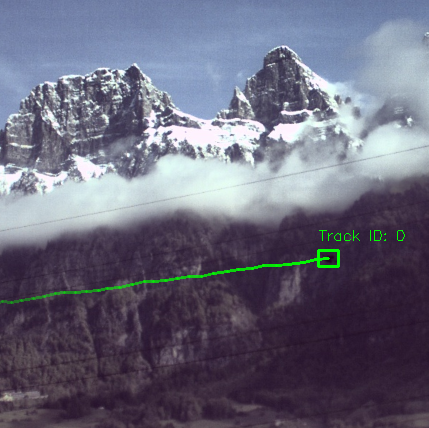}
     \end{subfigure}
     \hfill 
     \begin{subfigure}[b]{0.16\textwidth}
         \centering
         \caption*{Dark}
         \includegraphics[width=\textwidth, height=\textwidth, keepaspectratio=false]{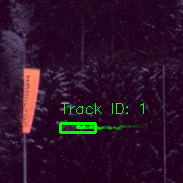}
     \end{subfigure}
     \hfill
     \begin{subfigure}[b]{0.16\textwidth}
         \centering
         \caption*{Groups}
         \includegraphics[width=\textwidth, height=\textwidth, keepaspectratio=false]{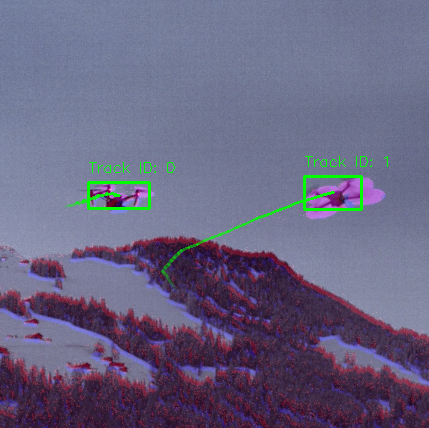}
     \end{subfigure}
     \hfill
     \begin{subfigure}[b]{0.16\textwidth}
         \centering
         \caption*{Sky bg.}
         \includegraphics[width=\textwidth, height=\textwidth, keepaspectratio=false]{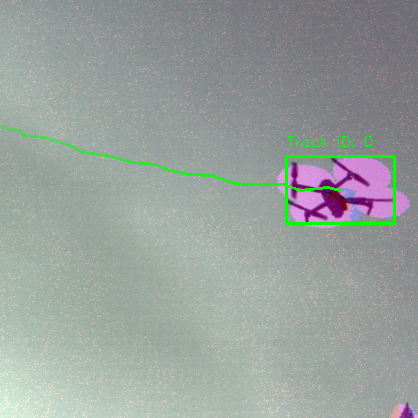}
     \end{subfigure}
     \hfill
     \begin{subfigure}[b]{0.16\textwidth}
         \centering
         \caption*{HDR}
         \includegraphics[width=\textwidth, height=\textwidth, keepaspectratio=false]{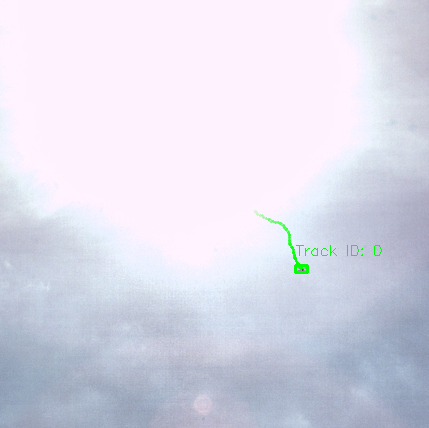}
     \end{subfigure}
     \hfill
     \begin{subfigure}[b]{0.16\textwidth}
         \centering
         \caption*{Urban}
         \includegraphics[width=\textwidth, height=\textwidth, keepaspectratio=false]{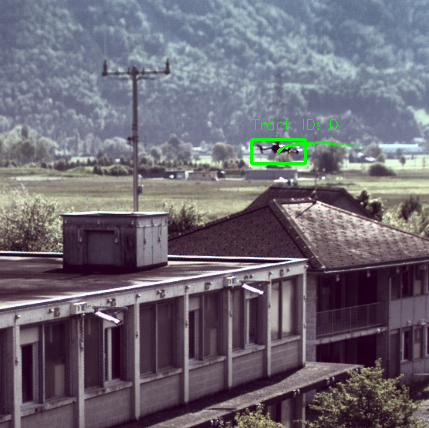}
     \end{subfigure}
     
     \caption{A sample of challenging scenarios from the dataset}
     \label{fig:sample_sequences}
\end{figure}
\vspace{-2em}

\subsection{Dataset Recording}
\begin{figure}[!htb]
  \centering
  \includegraphics{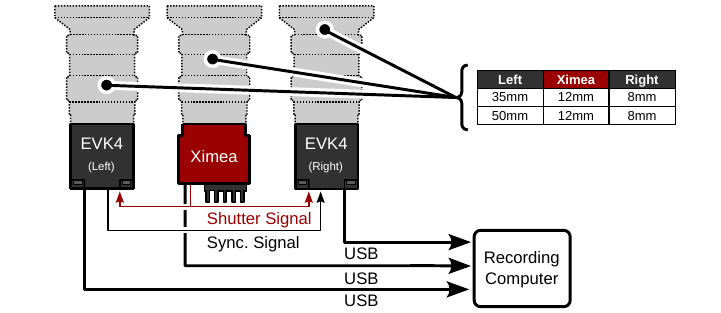}
  \caption{Schematic setup of the dataset recording setup, including lenses used on each camera. Global shutter RGB camera in center (Ximea) has its shutter signal hardware-timestamped using the EVK4 Trigger-in.} 
  \label{fig:dataset_recording_overview}
\end{figure}

The dataset was recorded using the multi-sensor setup shown in \cref{fig:dataset_recording_overview} and an assortment of drones listed in \cref{tab:drone_types}. The hardware setup consists of two event cameras (Prophesee EVK4) with C-mount lenses, as well as a conventional RGB frame camera (Ximea MQ022CG-CM-S7) mounted in parallel on a fixture. Following the multi-camera approach of \cite{2023moosmannQuantitativeEvaluationMultiModal}, this configuration allows for an efficient and accurate spatial mapping of labels from the RGB frames to both event sensors with varying magnification. However, this method results in a parallax effect, especially for nearby objects. The camera homographies were chosen to ensure that distant objects, such as the \gls{uav} targets, are aligned correctly. By using two event cameras with different \gls{fov} in parallel, we effectively double the data yield from a single recording and labeling pass. \\

\begin{center}
\begin{minipage}[t]{0.45\textwidth}
\vspace*{-\abovecaptionskip}
\captionof{table}{Drone types used for dataset collection with corresponding dimensions and maximum velocities. }
    \label{tab:drone_types}
    \resizebox{\linewidth}{!}{
    \begin{tabular}{lrrr}
    \toprule
    \textbf{Name} & 
\makecell[r]{\textbf{Size} \\ {[}mm{]}} & 
\makecell[r]{\textbf{Weight} \\ {[}g{]}} & 
\makecell[r]{\textbf{Speed} \\ {[}m/s{]}} \\
    \midrule
    Race Quad         & 220$\times$300                                     & 300                                & 33                                \\
    DJI FPV         & 120$\times$130                                     & 795                                & 26                                \\
    DJI Phantom 4 Pro & 280$\times$280                                        & 1388                               & 20                                \\
    DJI Mavic 2 Pro    & 322$\times$242                                        & 907                                & 20                                \\
    DJI Mini 3 Pro     & 251$\times$362                                        & 248                                & 15                                \\
    DJI Mini 4 Pro     & 298$\times$373                                        & 249                                & 16                                \\
    DJI Inspire 2     & 530$\times$605                                        & 3440                               & 23                                \\
    DJI Matrice 600   & 1668$\times$1518                                       & 9600                               & 18                   \\ 
    \bottomrule
    \end{tabular}
    }
\end{minipage}
\hfill
\begin{minipage}[t]{0.45\textwidth}
\vspace*{0pt}
    \captionof{listing}{Directory structure for labeled events and frames}
    \label{lst:dir_structure}
    \resizebox{\linewidth}{!}{%
        \begin{minipage}[t]{1.4\linewidth} 
        \vspace{0pt}
        \dirtree{%
            .1 sequence0.
            .2 events\_left.
            .3 000000.jpg.
            .2 events\_right.
            .3 000000.jpg.
            .2 events\_left.raw.
            .2 events\_right.raw.
            .2 frames.
            .3 000000.tiff.
            .2 frames\_ts.csv.
            .2 labels.txt.
            .2 h\_left.csv.
            .2 h\_right.csv.
            }
        \end{minipage}%
    }
\end{minipage}
\end{center}
The frame-based data was recorded at either 30FPS (to allow longer exposure times) or 60FPS (to capture higher temporal resolution). For different sequences in the recordings, a selection of low distortion (<0.12\%) lenses with focal lengths between \qty{8}{\milli\meter} and \qty{100}{\milli\meter} was used, with the exact configurations listed in the table of \cref{fig:dataset_recording_overview}. The cameras were all mounted on one tripod, with a disparity distance of \qty{5}{\centi\meter} between them. Data from all three cameras is transferred via USB connection to a recording computer. The data was recorded at as high a quality as possible without saturating the USB interface. In the case of the RGB camera, this was done by recording the full 10-bit RAW frames without demosaicing. The event cameras had the \gls{erc} set to \SI{100}{\mega Ev\per\second}. One of the key issues in high-quality sensor fusion of frame-based and event-based recordings is the need for highly precise time synchronization between the cameras. In our dataset, we achieve this by sharing the time-based synchronization signal between the two event cameras, as well as recording the frame camera's shutter signal as an external event directly into the event stream. This ensures that we can extract the exact point in time of the frame camera shutter in post-processing of the event camera streams.

\subsection{Dataset annotation protocol}
The long recordings are split into shorter $\sim$\SI{10}{\second} sequences of interest and aligned in the time domain using the hardware trigger events. Spatial alignment is achieved via a homography transformation. The homography matrix $H$ that maps a point $p_1$ in Camera 1 to a point $p_2$ in Camera 2 is given by \cref{eq:homography}. The homography can only be accurate for a single distance $d$, leading to parallax and misalignment. However, for large enough distances, $d$ becomes infinite, making the $\frac{t n^T}{d}$ term disappear.  The intrinsic matrices $K_1$ and $K_2$, as well as the rotation $R$, can be extracted using event-RGB camera calibration \cite{2021muglikarHowCalibrateYoura}.

\begin{equation}
    H = K_2 \left( R - \frac{t n^T}{d} \right) K_1^{-1}
    \label{eq:homography}
\end{equation}

\begin{figure}
    \centering
    \begin{minipage}{0.48\textwidth}
    \includegraphics[width=\linewidth]{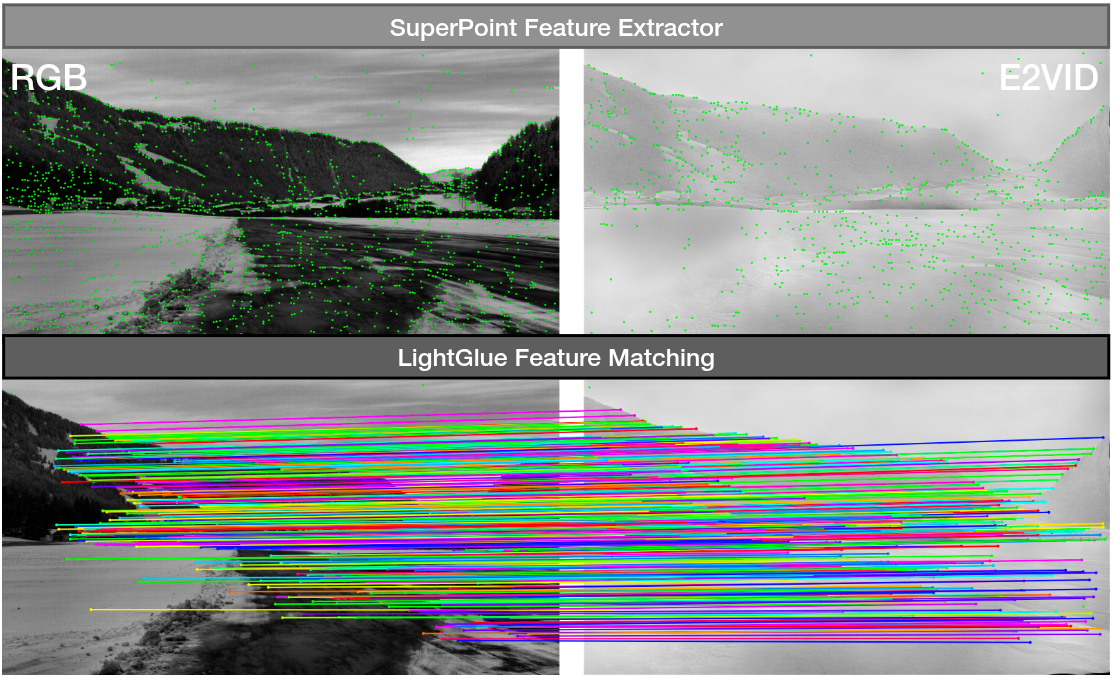}
    \caption{LightGlue Feature matching between RGB images and Synthetic frames from events using E2VID \cite{2021rebecqHighSpeedHigha}.}
    \label{fig:light_glue}
\end{minipage}\hfill
\begin{minipage}{0.48\textwidth}
    \centering
    \includegraphics[width=0.75\linewidth]{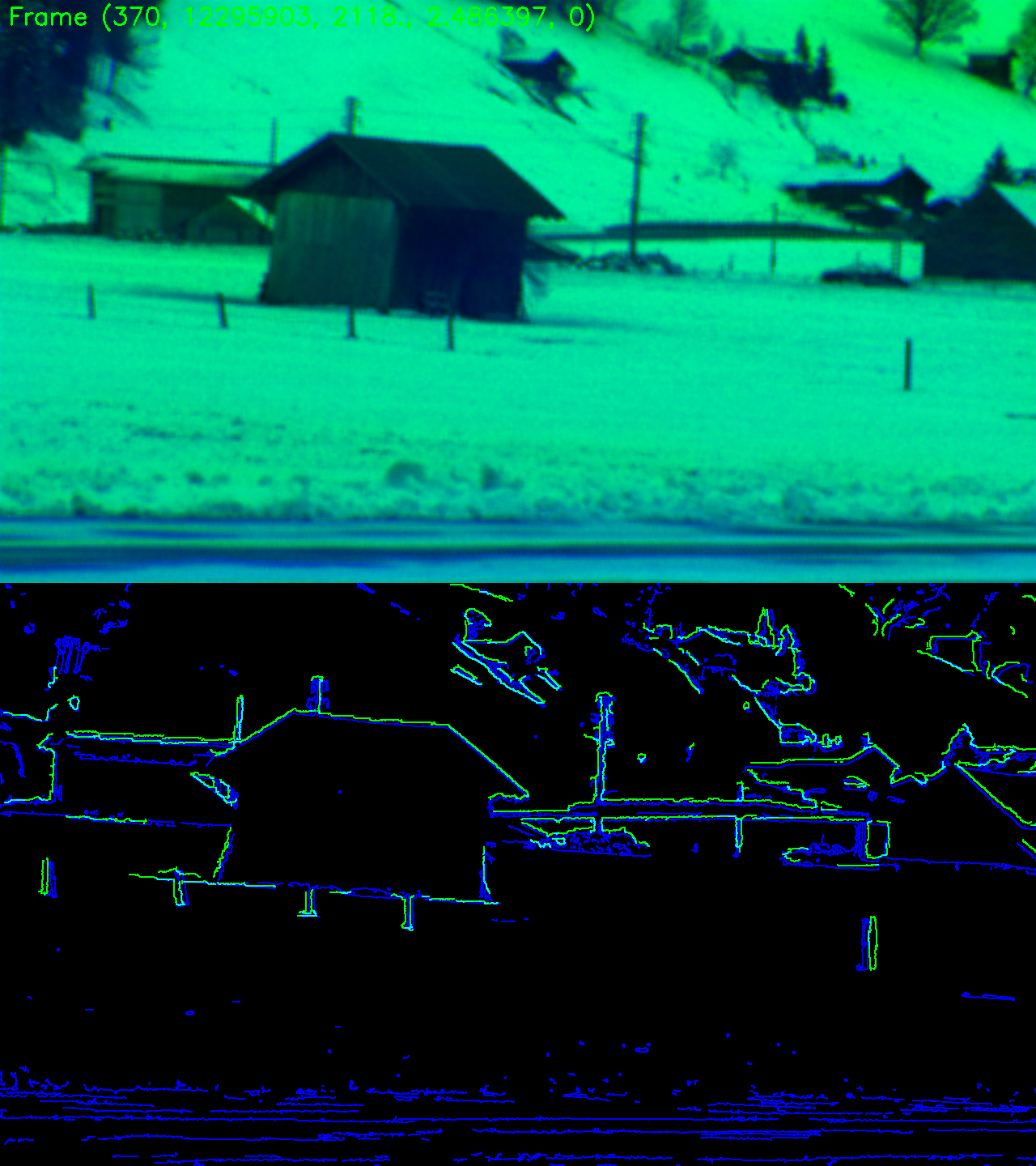}
    \caption{Validation of edge alignment by comparing the overlap of edges from a Canny edge detector \cite{1986cannyComputationalApproachEdge} (blue RGB, green events).}
    \label{fig:alignment}
\end{minipage}
\end{figure}

For some scenarios, spatial feature matching using LightGlue \cite{2023lindenbergerLightGlueLocalFeature} was utilized to recover the homography matrix as shown in \cref{fig:light_glue}. To verify the quality of image registration, edge overlap is compared between the modalities as shown in \cref{fig:alignment} and adjustments are made in cases where objects fall outside of the assumption that $d$ becomes infinite.

The event data was rendered as aggregated event-histogram frames. Together with the RGB-frames, these sequences were imported into Label Studio \cite{2020tkachenkoLabelStudioData} where the bounding boxes are manually labeled on the first frame (RGB or event), depending on which modality offers better object visibility. Robust Visual Tracking by Segmentation (RTS) \cite{2022paulRobustVisualTracking} was subsequently used to propagate the tracking across the entire sequence. Finally, any missing or incorrect bounding boxes were manually corrected. The label coordinates in one view are then automatically mapped to other views using the known homography transformation.

\subsection{Dataset Structure}
As shown in \cref{lst:dir_structure}, each sequence directory contains 10-bit demosaicked \cite{2021wangCompactHighQualityImage} frames in TIFF format together with aligned thumbnail Event-frames. The raw event data is stored in the native EVT3 format  \cite{2023MetavisionSDKMetavision} with the timestamps aligned to the frames. Relevant metadata, such as frame timestamps, exposure times, and gain, as well as the homography matrices, are recorded in CSV files, and the tracking bounding box labels are stored in the 2D MOT format \cite{2015lealtaixeMOTChallenge2015Benchmark}.

\subsection{Dataset Statistics}

\begin{figure}[h]
    \centering
    \includegraphics[width=\textwidth]{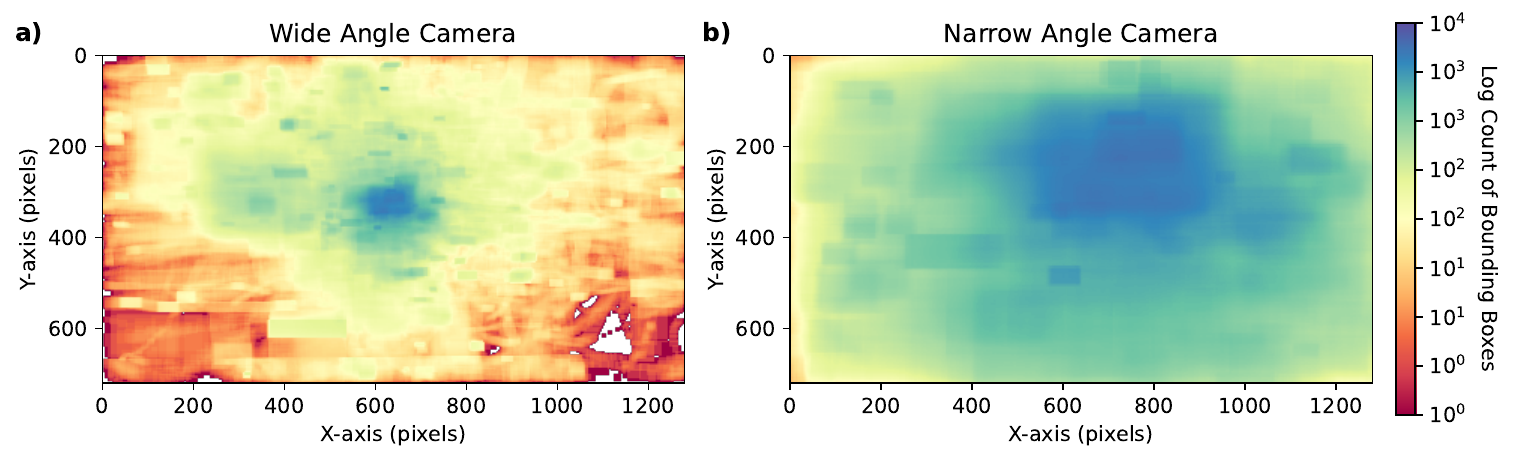}
    \caption{Bounding box heatmap for the dataset, split for Wide-angle event camera \gls{fov} (a) as well as zoomed Narrow-angle \gls{fov} (b). The centered distribution helps by allowing more aggressive dataset augmentations without loss of training labels.}
    \label{fig:bbox_heatmap}
\end{figure}

\begin{figure}[htbp]
    \centering
    \begin{minipage}{0.48\textwidth}
        \centering
        \includegraphics[width=\linewidth]{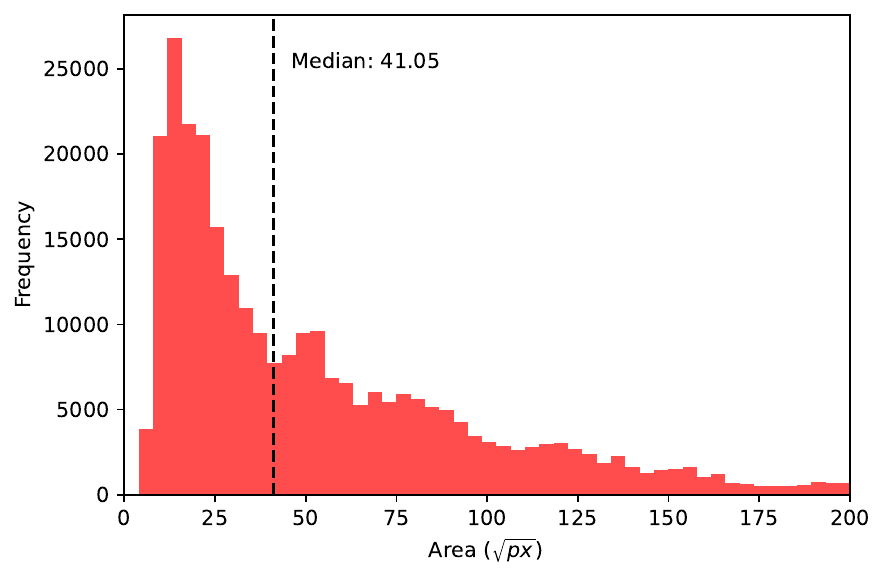}
        \caption{Distribution of bounding sizes in the SkyEV dataset. }
        \label{fig:areas}
    \end{minipage}\hfill
    \begin{minipage}{0.48\textwidth}
        \centering
        \includegraphics[width=\linewidth]{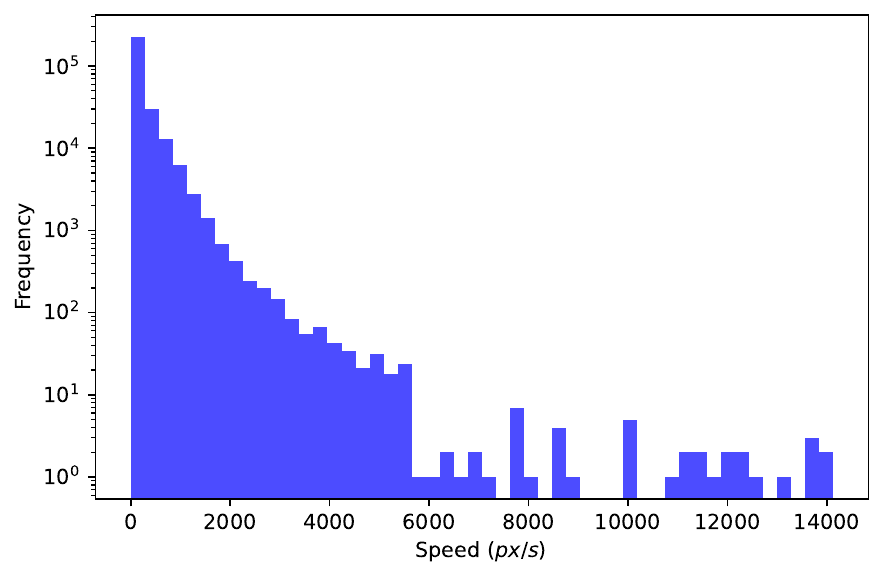}
        \caption{Distribution of bounding box velocities in the SkyEV dataset.}
        \label{fig:speeds}
    \end{minipage}
\end{figure}

The SkyEV dataset comprises 398 synchronized sequences, totaling 243,823 frames and 294,549 labeled bounding boxes, with an average of 1.2 bounding boxes per frame. Since the sequences were captured by 2 event-based cameras with differing \glspl{fov}, the dataset quantity is effectively doubled in the event domain. Even though the event-camera data is independent, the RGB frames are shared for these sequences (though with different homography projections). This is accounted for in the train/test split by dividing based on independent sequences.  As shown in the bounding box distribution heatmap in \cref{fig:bbox_heatmap}, the wide-angle camera contains some sparse areas. However, the combined dataset covers all pixels with a strong center bias. 
While dataset augmentation can mitigate the scarcity of small-scale bounding box instances to an extent, it often fails to capture the stochastic nature of sensor noise \cite{2008lichtsteiner128times128120}. The distribution of bounding box sizes in \cref{fig:areas} shows a bias towards boxes below 25 pixels with a mean $\sqrt{wh}$ of about 64px.  

Since \gls{uav} can move rapidly across the sensor \gls{fov}, the dataset was tailored to include high-velocity as well as high-resolution label (60FPS) bounding box targets. \cref{fig:speeds} shows the frequency distribution of tracking velocities. With the mean velocity of \SI{206.5}{\px\per\second}, maximum recorded velocity of \SI{14130}{\px\per\second}, and overall around 3\% of the tracks having a velocity \(>\SI{1000}{\px\per\second}\), the dataset provides a wide range of scenarios to benchmark detection and tracking models in these extreme scenarios.

\section{Experimental Baseline}
\begin{figure}[t]
  \centering
  \includegraphics{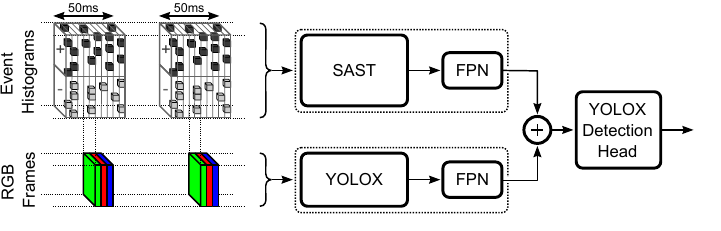}
  \caption{Overview of the SAST+RGB architecture used as experimental baseline.}
  \label{fig:sast_rgb_overview}
\end{figure}

Combined with our open-source dataset, we provide a state-of-the-art detection performance baseline, fusing both RGB and event camera data for drone object detection. The architecture's input consists of an RGB frame and a corresponding event histogram, with events sampled within a $\pm$\qty{25}{\milli\second} window centered on the RGB frame’s timestamp and divided into 10 temporal and 2 event polarity bins. The event histograms partially overlap between camera frames. The baseline architecture overview is shown in \cref{fig:sast_rgb_overview}. To satisfy the network's architectural constraints and maintain high performance on the target hardware, both RGB and event data are down-sampled from \qtyproduct[product-units=single]{1280x720}{\px} to \qtyproduct[product-units=single]{640x360}{\px}. This reduction reduces the computational load for real-time inference, with the input subsequently padded to a final resolution of \qtyproduct[product-units=single]{640x384}{\px} to meet the model's specific input requirements.\\


\begin{figure}[h]
    \centering
    \begin{minipage}{0.5\textwidth}
        \centering
        \includegraphics[width=\linewidth]{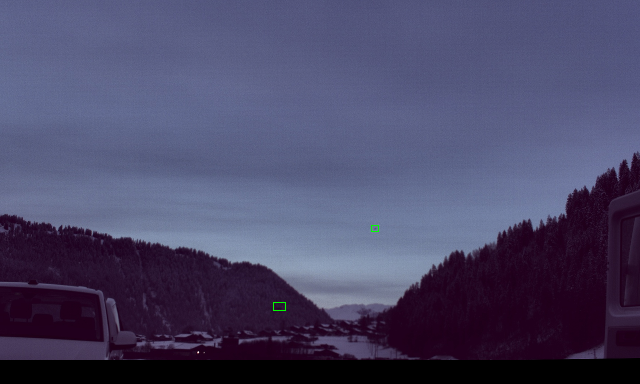}
    \end{minipage}\hfill
    \begin{minipage}{0.5\textwidth}
        \centering
        \includegraphics[width=\linewidth]{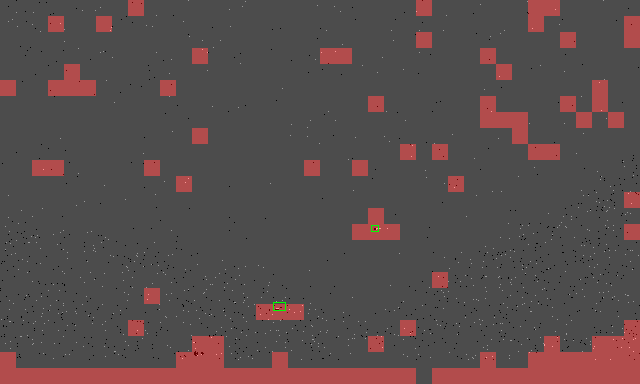}
    \end{minipage}
    \caption{The SAST model can selectively mask inactive event patches, thereby improving inference performance. In tasks such as \gls{uav} detection, large areas of the scene that lack activity allow the system to benefit greatly from this sparsity. }
    \label{fig:sast_vs_rgb}
\end{figure}

In terms of network architecture, only event histogram image tiles with potentially interesting features are extracted with an SAST \cite{SAST} module. In our approach, the SAST model was modified to accommodate unlabeled frames, addressing a limitation of the original architecture that required each sequence to contain at least one annotated frame. Additionally, the filtering of small bounding boxes \cite{2020perotLearningDetectObjects} \cite{YansongPeng2024SAST} in the evaluation process was removed. This is to accommodate our dataset, with most bounding boxes being in the small category \cite{2015linMicrosoftCOCOCommon} (Bounding boxes smaller than \qtyproduct[product-units=single]{32x32}{\px}). Otherwise, the evaluation of the SAST implementation would filter those boxes. \\

The SAST module is combined with a YOLOX \cite{2021geYOLOXExceedingYOLO} backbone that takes RGB images as input. 
To extract the same amount of semantic information from RGB data as SAST does from event data, YOLOX is used as the backbone for the RGB stream. After feature extraction, the outputs of both networks are aligned at the \gls{fpn} stage and combined, ensuring that information from the two modalities is fused at a comparable semantic level. 
Alternative approaches using smaller RGB backbones yielded degraded performance, as the high information density of the RGB stream tended to overshadow the event data, while the backbone's reduced capacity limited its ability to extract sufficient features. \\

The features from the RGB and event streams are fused at the \gls{fpn} stage. Let $P_i^{\text{RGB}}$ and $P_i^{\text{Ev}}$ represent the hierarchical feature maps extracted by the YOLOX and SAST \gls{fpn} stages, respectively, where $i$ denotes the layers hierarchy. The fused feature pyramid $P_i^{\text{Fused}}$ is obtained through an element-wise summation across corresponding scales:

\begin{equation}
    P_i^{\text{Fused}} = P_i^{\text{RGB}} \oplus P_i^{\text{Ev}}, \quad \text{for } i \in \{1, 2, 3\}
\end{equation}
where $P_i \in \mathbb{R}^{C_i \times H_i \times W_i}$. This additive fusion ensures that the high-resolution spatial details from the RGB stream are directly augmented by the sparse, motion-aware tokens generated by the SAST module at every semantic level. The final detection is done by a YOLOX detection head on the combined hierarchical extracted features $P_i^{\text{Fused}}$. \\

This design allows each modality to leverage a strong, specialized architecture while enabling effective multimodal integration. 
Furthermore, this specifically enables a direct comparison of the influence of event data on overall performance, since both modules, SAST and YOLOX, can be run independently and evaluated separately.

\section{Baseline Results}

Performance was benchmarked independently across individual modalities (frame and event data) as well as the integrated fusion model. YOLOX was tested on the exclusive RGB frames and event frames rendered from the asynchronous data as event histograms \cite{parkContinuousHistogramEventbased}. 

An 80:20 split for training and evaluation was used, and the following data augmentation was applied: 50\% chance of horizontal flips and 80\% probability of zoom. Zoom-in was configured to be twice as likely as zoom-out. The zoom factors for zoom-in and zoom-out were chosen to be 1.5 and 1.2, respectively. Each model on the datasets was trained for 20 epochs, and the best mAP$_{50-95}$ along with its corresponding AP$_{\text{small}}$ are displayed in \cref{tab:results}. The AP$_{\text{small}}$ score is evaluated on bounding boxes with size smaller than $30\times30$ as defined in \cite{2015linMicrosoftCOCOCommon}.

\begin{table}[ht]
\caption{Benchmark results comparing RGB and Event-based models across hardware. RTX4090 inference was used using CudaTimer in PyTorch. On Jetson AGX Orin, the models were optimized using TensorRT. }
\centering
\setlength{\tabcolsep}{4pt} 
\begin{tabular}{@{} l *{6}{c@{\extracolsep{2mm}}} @{}}
\toprule
\textbf{Method} & 
\textbf{AP}$_{\text{\tiny{small}}}$ & 
\textbf{mAP}$_{\text{\tiny{50-95}}}$ & 
\textbf{Params } &
\textbf{Compute}& 
\textbf{RTX4090} & 
\textbf{AGX Orin} \\
 & & & \text{[M]} & [{GFLOPs}] & \text{[ms]} & \text{[ms]} \\
\midrule
YOLOX-RGB    & 32.9 & 35.5 & 9.0  & 16.2 & \textbf{2.61} & \textbf{1.92} \\
YOLOX-Events & 32.2 & 37.3 & 9.0  & 18.6 & --   & 3.64 \\
SAST         & 40.7 & 43.4 & 18.9 & 42.0 & 6.55 & 4.98 \\
SAST+RGB   & \textbf{42.3} & \textbf{45.8} & 31.3 & 52.3 & 7.92 & 5.26 \\
\bottomrule
\end{tabular}
\label{tab:results}
\end{table}

Comparing YOLOX running exclusively on the RGB or Event frames to a fuse model shows a significant improvement in mAP accuracy. As expected, filtering the validation for small bounding boxes shows a clear deterioration in performance. Regarding computational requirements, all models can achieve inference latency under \SI{6}{\milli\second} with the optimized TensorRT engine on embedded hardware such as the Jetson AGX Orin. This allows an inference rate exceeding \SI{160}{\Hz}. This is significantly higher than the temporal resolution of the RGB frames, yet it can still take advantage of the high-resolution event-based data.

\section{Conclusion}
This paper provides the SkyEV open-source drone detection dataset, recording sequences with a total of \qty{2.17}{\hour} from two highly synchronized event-based cameras and one RGB frame camera, and ensures careful, precise labeling. Among these recordings, a diverse set of 8 different drone types is represented. The dataset provides labeled bounding box tracks in the MOT format for a diverse range of sizes, with a median bounding box size of 41 pixels and velocities of up to 14 000 pixels per second, leveraging the extreme speed of the event-based sensors. Next to the dataset, we presented and evaluated a state-of-the-art detection baseline, effectively fusing RGB frame data with event histograms. This SAST+RGB baseline is optimized towards efficient computing by ensuring that only potentially relevant histogram patches are considered for processing by a first SAST algorithm stage.

Combined, this work provides the groundwork for training and evaluation of efficient and high-performance drone detection algorithms.


%
%
\bibliographystyle{splncs04}
\bibliography{main}
\end{document}